\PassOptionsToPackage{square,comma,numbers,sort&compress}{natbib}  
\documentclass{article}

\usepackage[preprint]{neurips_2024}

\usepackage{abbrv}

\usepackage[utf8]{inputenc} 
\usepackage[T1]{fontenc}    
\usepackage[colorlinks]{hyperref} 
\usepackage{url}            
\usepackage{booktabs}       
\usepackage{amsfonts}       
\usepackage{nicefrac}       
\usepackage{microtype}      
\usepackage{xcolor}         
\usepackage{graphicx}
\usepackage{multirow}
\usepackage{float}
\usepackage{amsmath}
\usepackage{amssymb}
\usepackage{wrapfig}
\usepackage{subfig}
\usepackage{array}
\usepackage{lipsum}
\usepackage{amssymb}
\usepackage{pifont}
\usepackage[misc]{ifsym}

\newcommand{\xmark}{\ding{55}}

\usepackage{colortbl}
\usepackage{tabulary,multirow,overpic,xcolor}
\definecolor{mygray}{gray}{0.92}
\definecolor{baselinecolor}{gray}{.9}

\newlength\savedwidth

\newlength\savewidth

\makeatletter
\def\blfootnote{\xdef\@thefnmark{}\@footnotetext}
\makeatother

\title{Open-Vocabulary Spatio-Temporal Action Detection}

\author{{Tao Wu\textsuperscript{1} \quad Shuqiu Ge\textsuperscript{1} \quad Jie Qin\textsuperscript{2} \quad Gangshan Wu\textsuperscript{1} \quad Limin Wang\textsuperscript{1,3,~\Letter}} \\ 
$^1$State Key Laboratory for Novel Software Technology, Nanjing University \\
$^2$Nanjing University of Aeronautics and Astronautics \quad $^3$Shanghai AI Lab \\\\
}

\begin{document}

\maketitle

\begin{abstract}
Spatio-temporal action detection (STAD) is an important fine-grained video understanding task. Current methods require box and label supervision for all action classes in advance. However, in real-world applications, it is very likely to come across new action classes not seen in training because the action category space is large and hard to enumerate. Also, the cost of data annotation and model training for new classes is extremely high for traditional methods, as we need to perform detailed box annotations and re-train the whole network from scratch. In this paper, we propose a new challenging setting by performing open-vocabulary STAD to better mimic the situation of action detection in an open world. Open-vocabulary spatio-temporal action detection (OV-STAD) requires training a model on a limited set of \textit{base} classes with box and label supervision, which is expected to yield good generalization performance on \textit{novel} action classes. For OV-STAD, we build two benchmarks based on the existing STAD datasets and propose a simple but effective method based on pretrained video-language models (VLM). To better adapt the holistic VLM for the fine-grained action detection task, we carefully fine-tune it on the localized video region-text pairs. This customized fine-tuning endows the VLM with better motion understanding, thus contributing to a more accurate alignment between video regions and texts. Local region feature and global video feature fusion before alignment is adopted to further improve the action detection performance by providing global context. Our method achieves a promising performance on \textit{novel} classes.  
\end{abstract}

\section{Introduction}
\label{sec:intro}
Spatio-temporal action detection (STAD) aims to localize action instances with bounding boxes in video frames and recognize their categories. It has wide applications in fields like security monitoring and sports video analysis. The current mainstream STAD methods include two-stage methods~\cite{lfb,feichtenhofer2019slowfast,aia,acarn,hit,yowo,acrn,vat,chen2021watch} which decouple human detection and action classification and query-based one-stage methods~\cite{zhao2022tuber,wu2023stmixer}. Despite remarkable progress in recent years, current methods require human bounding boxes and action labels for all action categories of interest for supervised training, and the trained model can only cope with the action classes seen during training.
\begin{figure}
\centering
\includegraphics[width=0.85\textwidth]{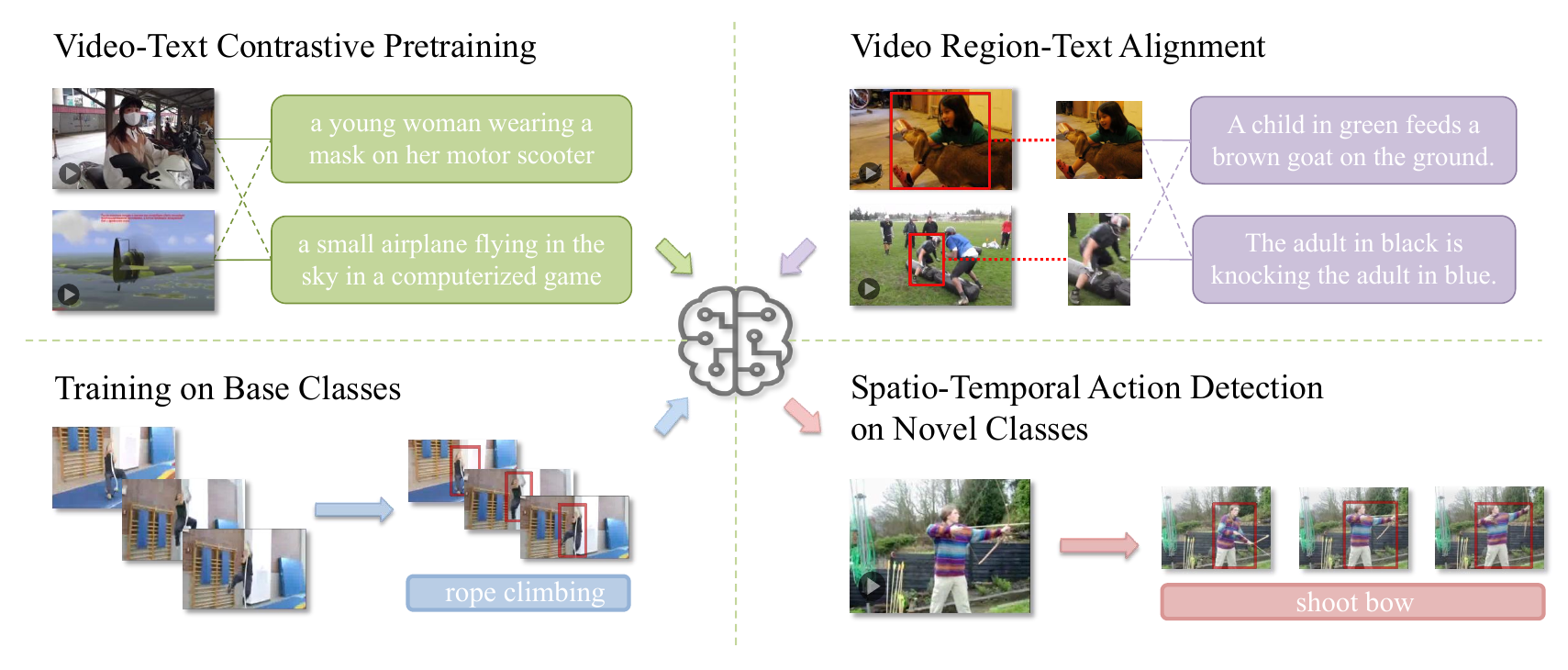}
\vspace{-10pt}
\caption{An overview of the open-vocabulary spatio-temporal action detection framework. We use a video-language model pretrained on massive video-text data as the video and text encoders. The model is further trained on video region-text pairs and finetuned on base classes. The trained model is expected to have a good generalization performance on novel action classes.  }
\label{OV-STAD}
\vspace{-20pt}
\end{figure}

Some vision-language models (VLMs), such as CLIP~\cite{radford2021learning} and ALIGN~\cite{jia2021scaling} have developed the ability of zero-shot visual recognition through contrastive pretraining on large-scale image-text datasets. Recently, some object detection methods~\cite{gu2021open,wu2023aligning,zhong2022regionclip,zang2022open,kuo2022f} have exploited this ability of the VLMs for open-vocabulary object detection~\cite{zareian2021open}. They use the visual and text representations provided by the pretrained VLMs to train on a limited set of base classes for the target object detection task. The trained model is expected to achieve good generalization performance on novel object classes that have not been seen during the training process. As for the video domain, video-text data pretrained models like VideoCLIP~\cite{xu2021videoclip} and ViCLIP~\cite{wang2023internvid} have also made it possible to develop open-vocabulary methods for video recognition and detection tasks. There have been some works focusing on open-vocabulary action recognition~\cite{pan2022st,ma2022x,wang2021actionclip,rasheed2023fine,momeni2023verbs,jia2023generating} or temporal action detection~\cite{ju2022prompting,rathod2022open}, but to the best of our knowledge, more fine-grained STAD task has not yet been studied under the open-vocabulary setting.  

However, we believe that open-vocabulary spatio-temporal action detection (OV-STAD) is of great research value. The category space of actions is huge and it is tough to enumerate all action categories when building a spatio-temporal action detection dataset. Also, the labeling cost of STAD data is high because it requires fine-grained annotation, that is, using bounding boxes to accurately mark all occurrences of action instances in each frame of the video and tag them with category labels. In real-world applications, it is very likely to encounter action categories that are not included in existing spatio-temporal detection datasets. Every time new action categories are encountered, it will consume a lot of resources to re-collect and label the data of the new categories and re-train the model. Under the open-vocabulary setting, spatio-temporal action detection models are trained on a limited set of action classes with action boxes and labels provided for supervision and the trained models are expected to have good generalization performance on unseen action categories, which can effectively reduce data annotation costs and improve model training efficiency. 

OV-STAD is challenging in at least three aspects: First, there is no existing benchmark for model training and evaluation. Second, there is a gap between the upstream pretraining and the target action detection task. The upstream pretraining aligns texts with global video features, while the target task requires alignment between action labels and local video patches. Besides, in pretraining, a large proportion of texts only involve scene, object, and appearance descriptions, but are not related to people's actions or activities. Therefore, the pretrained model has certain deficiencies in understanding motion concepts~\cite{momeni2023verbs,wang2024paxion}. Third, the number of action categories in the current spatio-temporal action detection datasets is relatively small. During the training process, the model tends to overfit the base classes.

In this paper, we deal with the problem of OV-STAD. We construct two benchmarks for it based on UCF101-24~\cite{ucf101}, JHMDB-21~\cite{jhmdb} and AVA~\cite{ava} by properly splitting action classes into novel and base classes. We test some video-language models on the two benchmarks. Directly applying these models to the OV-STAD task performed poorly, although they achieved good results on zero-shot action recognition. We attribute this to the gap between the upstream pre-training and the target task as mentioned above. As shown in Figure~\ref{OV-STAD}, to close this gap, we propose to use video region-text pairs to enhance the alignment between texts and local video patches. The video region-text pairs are selected from spatio-temporal grounding datasets~\cite{vidstg,hcvg}. They are all human-centered, and the texts contain high-density descriptions of human actions and activities. The model can learn rich motion concepts intensively. Besides, we propose to fuse the global video-level feature and the local patch-level feature before aligning with the action prompts, which retains knowledge from large-scale video-text pretraining to alleviate overfitting and also provides a global context for action recognition. We conduct extensive experiments on the two benchmarks to demonstrate the effectiveness of our designs. Our method has achieved promising results on novel action classes. In summary, our contribution is threefold: 1) We propose the new open-vocabulary setting for the STAD task and establish two benchmarks for model training and evaluation based on existing datasets. 2) We propose a simple but effective method based on video-language models, which leverage video region-text alignment and global-local fusion to close the gap between upstream pretraining and the target action detection task. 3) We conduct extensive experiments on the two benchmarks to demonstrate the effectiveness of our designs and our method achieves promising results on novel classes.

\section{Related Work}

\noindent\textbf{Open-Vocabulary Object Detection.} Open-vocabulary object detection (OVOD)~\cite{zareian2021open} aims to detect objects of new categories whose annotations are not provided during the training process. F-VLM~\cite{kuo2022f} uses frozen pretrained VLM encoders for feature extraction, training an additional detection head. Some works~\cite{gu2021open,wu2023aligning,zhong2022regionclip,zang2022open} align regional embeddings with the representation space of CLIP via knowledge distillation. Others~\cite{yao2022detclip,feng2022promptdet,bangalath2022bridging,li2022grounded,zhong2022regionclip,zhao2022exploiting,zhou2022detecting,gao2021towards} generate pseudo object labels or text descriptions for image regions using VLMs. The generated data are used to train the detector for learning diverse object category concepts. GLIP~\cite{li2022grounded} and GroundingDINO~\cite{liu2023grounding} identify the correlation between object detection and visual grounding tasks ~\cite{kamath2021mdetr,krishna2017visual,kazemzadeh2014referitgame}, thus leveraging visual grounding datasets for visual concept learning.

\noindent\textbf{Open-Vocabulary Action Recognition.} Early OVAR methods~\cite{pan2022st,ma2022x,wang2021actionclip,rasheed2023fine} adapt image-text pretrained VLMs into action recognition models by adding additional learnable adapters and temporal modeling layers.~\cite{momeni2023verbs,jia2023generating} leverage large language models (LLMs)~\cite{chowdhery2023palm,achiam2023gpt} to craft diverse natural language descriptions for videos. ~\cite{momeni2023verbs} strengthens the model's ability to recognize actions by constructing hard negatives through the substitution of verbs in captions. VideoCLIP~\cite{xu2021videoclip} utilizes instructional video datasets and InternVid~\cite{wang2023internvid} employs LLMs to construct a large-scale video-text dataset, pretraining a CLIP-like model suitable for the video field to bridge the gap between image and video modalities.

\noindent\textbf{Spatio-Temporal Action Detection}
Spatio-temporal action detectors can be categorized into and two-stage ones~\cite{fan2021multiscale,feichtenhofer2019slowfast,lfb,tong2022videomae,acarn,aia,chen2021watch,vat,yowo,acrn} and one-stage ones~\cite{li2020actions,zhao2022tuber,wu2023stmixer}. Two-stage methods decouple the STAD task into two steps of localization and classification. Some works~\cite{fan2021multiscale,feichtenhofer2019slowfast,lfb,tong2022videomae,acarn,aia} utilize extra human detectors~\cite{fasterrcnn} to detect human bounding boxes in video keyframes. They then extract features for each human box through ROI operations~\cite{girshick2015fast,he2017mask} for action classification. Other works~\cite{chen2021watch,vat,yowo,acrn} jointly train the human proposal generation network and classification network, achieving an end-to-end detection process. One-stage methods perform localization and classification simultaneously. MOC~\cite{li2020actions} densely regresses bounding boxes and predicts action classification scores for each point on the feature map. Inspired by DETR~\cite{carion2020end}, TubeR~\cite{zhao2022tuber} employs a set of tubelet queries to directly generate action tubes on input video clips. STMixer~\cite{wu2023stmixer} constructs a multi-scale spatio-temporal feature space and flexibly samples and decodes features from it.

\section{Method}

\begin{figure}
\centering
\includegraphics[width=\textwidth]{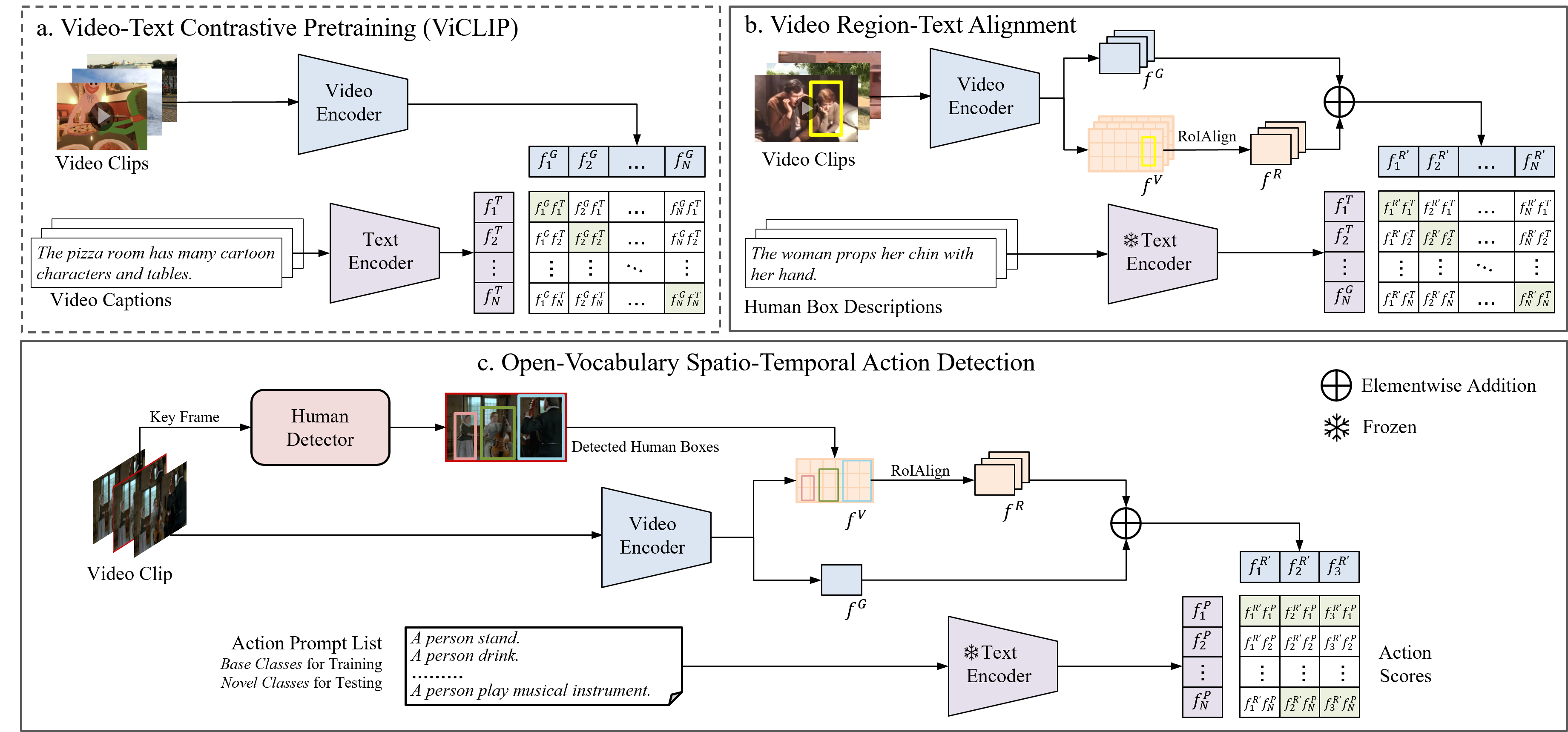}
\caption{The pipeline of the proposed OV-STAD method. A video-language model pretrained on large-scale video-text datasets is used for image and text feature extraction to leverage its zero-shot visual recognition ability. We further train the model on video region-text pairs to acquire a better alignment between video regions and texts. Finally, for OV-STAD, a two-stage detection pipeline is adopted. A human detector is first employed to generate human proposals on the keyframes and then action classes are recognized by aligning the region features to action prompt embeddings.}
\label{pipeline}
\vspace{-15pt}
\end{figure}

In this section, we present the proposed OV-STAD method, as illustrated in Figure~\ref{pipeline}. We adopt a VLM pretrained on large-scale video-text datasets for visual and textual feature extraction to leverage its capability of understanding diverse visual concepts. A two-stage detection pipeline is employed, which detects human bounding boxes first and then extracts features for each box to align with action prompt embeddings for action classification. However, local visual features are deficient in motion representation and are not well-aligned with text descriptions. We propose to enhance video region features through global and local video feature fusion, as well as video region-text alignment.

\subsection{Revisiting Video-Text Contrastive Pretraining}

We use the ViCLIP model~\cite{wang2023internvid} pretrained on the InternVid-10M-FLT dataset for video and text feature extraction. The InternVid-10M-FLT dataset contains 10M video-text pairs. ViCLIP learns video-text representations by optimizing video-text alignment. Specifically, as shown in Figure~\ref{pipeline}(a), ViCLIP employs a video encoder and a text encoder to acquire global video features $f^\mathbf{G}$ and text features $f^\mathbf{T}$.To improve the similarity between the video feature and its corresponding text feature while reducing its similarity with unmatched text features, ViCLIP calculates and minimizes the InfoNCE loss based on $f^\mathbf{G}$ and $f^\mathbf{T}$, given by:
\begin{equation}
\mathcal{L} =-\sum_{i=1}^{N} \log \frac{\exp \left(\operatorname{sim}\left(f_{i}^{\mathbf{G}}, f_{i}^{\mathbf{T}}\right) / \tau\right)}{\sum_{j=1}^{N} \exp \left(\operatorname{sim}\left(f_{i}^{\mathbf{G}}, f_{j}^{\mathbf{T}}\right) / \tau\right)},
\end{equation}
where $N$ denotes the number of video-text pairs in the batch, sim$(\cdot)$ computes cosine similarity between two features and $\tau$ is a learnable temperature.

\subsection{Two-Stage OV-STAD Pipeline}

As shown in Figure~\ref{pipeline}(c), we adopt a two-stage detection pipeline for OV-STAD. We first use an offline action-agnostic human detector to generate human proposals on the keyframes and then recognize the action of each proposal. We finetune the model on the base classes by aligning the features of the human proposals with the action category prompts. Specifically, video clips centered at the keyframes are input to the video encoder to extract global video features $f^\mathbf{G}$ and video feature maps $f^\mathbf{V}$. Local region features $f^\mathbf{R}$ are acquired by applying RoIAlign~\cite{he2017mask} on $f^\mathbf{V}$ with human box proposals $\mathbf{B}$ detected on the keyframes.  For the text branch, a simple prompt template of ``a person [\textit{action class}]'' is used. The text encoder is employed to extract prompt embeddings $f^\mathbf{P}$. To preserve the video-text alignment learned by pretraining on massive video-text data and also provide a global context for local features, we propose to fuse the global video features $f^\mathbf{G}$ and the local region features $f^\mathbf{R}$ by simple element-wise addition, that is,
\begin{equation}
    f^{\mathbf{R}^{\prime}} =\beta \cdot f^\mathbf{R} + (1-\beta) \cdot f^\mathbf{G},
\end{equation}
where $\beta$ is a preset hyperparameter used to control the proportion of local and global information. We calculate the cosine similarity between the fused video features $f^{\mathbf{R}^{\prime}}$ and action prompt embeddings $f^\mathbf{P}$ as action classification scores $S$, given by:
\begin{equation}
    S_{ij} = \operatorname{sigmoid}(\operatorname{sim}(f_{i}^{\mathbf{R}^{\prime}}, f_{j}^{\mathbf{P}}) / \tau),
\end{equation}
where $Sij$ is the score of the $i$-th human box on the $j$-th action category.
For training, the prompts of the base action classes are provided, and a focal loss~\cite{focalloss} is calculated based on the output scores $S$ and the ground-truth $Y$, given by:
\begin{equation}
    \mathcal{L}=-\sum_{i=1}^{N}\sum_{j=1}^{C_b}\left\{\begin{array}{lc}
\left(1-S_{ij}\right)^{\gamma} \log S_{ij} & \text { if } j=Y_i \\
S_{ij}^{\gamma} \log \left(1-S_{ij}\right) & \text { otherwise }
\end{array}\right.
\end{equation}
where $N$ denotes the number of human proposals in the batch, and $C_b$ denotes the number of base classes. For inference, the action prompts should be replaced with action prompts of interest, which can include both the base classes and novel classes that have not been seen during the training process.

\subsection{Video Region-Text Alignment}

Although the VLM has gained the ability to associate videos with their corresponding text descriptions after contrastive pretraining on massive video-text data, we argue that there is still a gap between the upstream pretraining and the target action detection task: First, in pretraining, global video features and text features are aligned, while the target action detection task requires aligning local video patches (i.e human bounding boxes) with action category prompts. Second, a large proportion of videos or text descriptions only involve scenes, objects, and appearances, but are not related to people's actions or activities.  Studies~\cite{momeni2023verbs,wang2024paxion} have demonstrated that these pretrained models are not sufficient for understanding diverse motion concepts.

To better adapt the holistic VLM for the fine-grained action detection task, we propose to further train the model on video region-text data for better alignment between video regions and their corresponding text descriptions. We collect video region-text pairs from spatio-temporal video grounding (STVG) datasets, including HC-STVG2.0~\cite{hcvg} and VidSTG~\cite{vidstg}. The STVG task requires localizing the target object described by the given text with a spatio-temporal tube (\ie a sequence of bounding boxes). HC-STVG2.0 contains 10,131 samples for training and 3,482 samples for validation. The text descriptions in HC-STVG2.0 are all about human actions and activities, but the descriptions usually contain multiple sentences describing a series of consecutive actions. It may introduce lots of noise to align a human box whose visual representation is extracted by applying RoIAlign~\cite{he2017mask} on feature maps of the 8-frame-long video clip centered at it with the whole text. To reduce noise in region-text alignment, we break long sentences in descriptions into short sentences, each containing only one action or activity. At the same time, we segment the tubes into sub-tubes and match them one-to-one with the short sentences in chronological order. VidSTG contains 99,943 video-sentence pairs. We selected 23,165 samples related to human actions and activities with declarative sentences from it. The text descriptions in VidSTG are all short sentences. We attach the sentence to each bounding box of the corresponding tube to form video region-text pairs. More details about data processing can be found in the \textit{Appendix}.

\begin{figure}[t]
\centering
\includegraphics[width=0.9\textwidth]{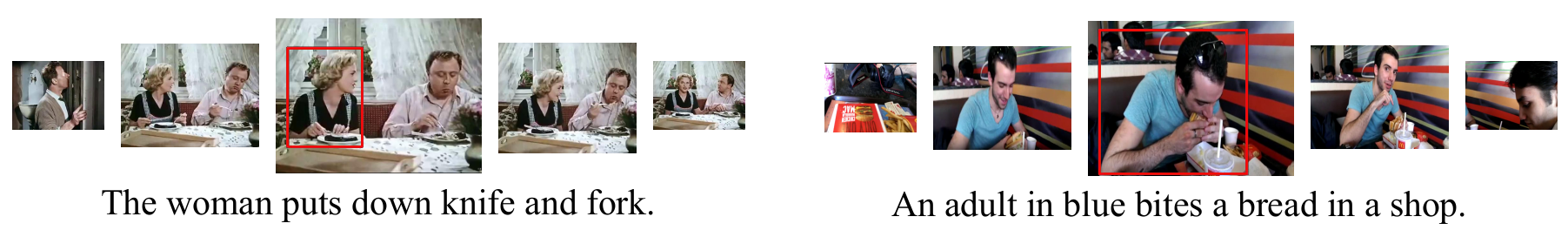}
\vspace{-5pt}
\caption{Two sample video region-text pairs from the HC-STVG2.0~\cite{hcvg} (left) and VidSTG~\cite{vidstg} (right) datasets. The text descriptions focus on human actions and activities. }
\label{sample}
\vspace{-15pt}
\end{figure}

As shown in Table~\ref{tab:stat}, this resulting dataset contains 7,066,136 human bounding box-text pairs. Since one text description can correspond to multiple boxes in consecutive frames referring to the same person, there are a total of 36,778 different text descriptions. We show 2 data samples in Figure~\ref{sample}. Different from the previous video-text pretraining data, the text descriptions are all about human actions and activities. In the process of aligning video regions and text descriptions, the model is exposed to various action descriptions and learns to distinguish diverse motion concepts. 

\begin{table}[t]
\centering
\setlength{\tabcolsep}{10pt}
\scriptsize
\caption{Statistics of the dataset for video region-text alignment.}
\label{tab:stat}
\begin{tabular}{l|ccc}
\toprule
 & \#box-text pairs & \#unique sentences & \#avg. boxes per sentence \\ \midrule
from VidSTG~\cite{vidstg} & 5,182,090 & 23,165 & 223.7 \\
from HC-STVG2.0~\cite{hcvg} & 1,884,046 & 13,613 & 138.4 \\
ALL & 7,066,136 & 36,778 & 192.1 \\ \bottomrule
\end{tabular}
\vspace{-6pt}
\end{table}

The process of region-text alignment is depicted in Figure~\ref{pipeline}(b). Video-text data pretrained VLM is adopted for video and text feature extraction. The text encoder is frozen and we only train the video encoder in this stage as we aim for better video region representations. Given the video feature maps $f^\mathbf{V}$ and the human bounding boxes $\mathbf{B}$, the features $f^\mathbf{R}$ of the regions $\mathbf{B}$ are extracted by applying the RoIAlign operation~\cite{he2017mask} on $f^\mathbf{V}$. Global and local video feature fusion is also adopted the fused feature $f^{\mathbf{R}^{\prime}}$ is used for alignment with text embeddings. Focal loss~\cite{focalloss} is adopted to enhance the learning of hard samples. Formally,
\begin{equation}
    P_{ij} = \operatorname{sigmoid}(\operatorname{sim}(f_{i}^{\mathbf{R}^{\prime}}, f_{j}^{\mathbf{T}}) / \tau)
\end{equation}
\begin{equation}
    \mathcal{L}=-\sum_{i=1}^{N}\sum_{j=1}^{N}\left\{\begin{array}{lc}
\left(1-P_{ij}\right)^{\gamma} \log P_{ij} & \text { if } i=j \\
P_{ij}^{\gamma} \log \left(1-P_{ij}\right) & \text { otherwise }
\end{array}\right.
\end{equation}
where $N$ denotes the number of video region-text pairs in the batch, sim$(\cdot)$ computes cosine similarity between two features, $\tau$ is a learnable temperature, and $\gamma$ is a hyperparameter of the focal loss.

\begin{figure}[t]
\centering
\includegraphics[width=0.93\textwidth]{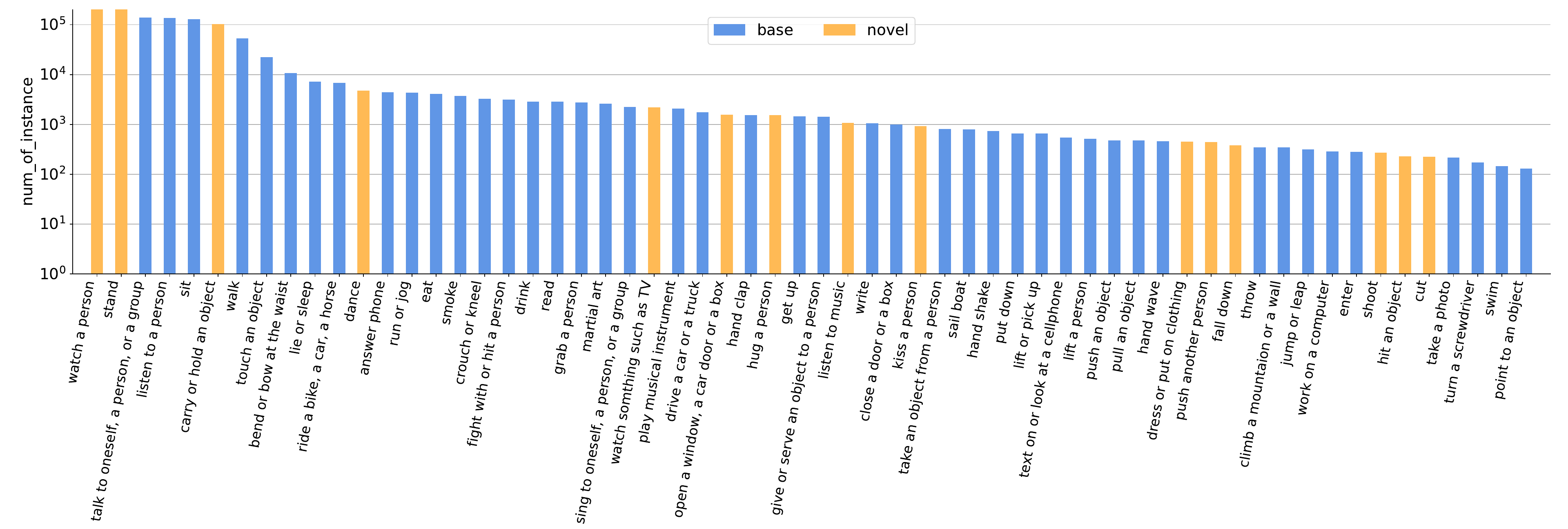}
\vspace{-5pt}
\caption{Division of base classes and novel classes on AVA~\cite{ava} for OV-AVA. Displayed in order of number of instances from high to low.}
\label{ava}
\vspace{-15pt}
\end{figure}
\subsection{Benchmark and Evaluation Metric}

For the newly proposed OV-STAD task, we propose two benchmarks for model training and evaluation based on the existing STAD datasets. 

\noindent \bf UCF-JHMDB. \rm UCF-JHMDB is based on UCF101-24~\cite{ucf101} and JHMDB-21~\cite{jhmdb}. The UCF101-24 dataset consists of 3207 videos annotated with action instances of 24 action classes. The action classes of UCF101-24 are mainly sports, such as ``\textit{volleyball spiking}''. The JHMDB-21 dataset consists of 928 videos from 21 action classes, including sports like ``\textit{swing baseball}'', daily activities and actions like ``\textit{brush hair}''. The 24 action classes from UCF101-24 are used as base classes, and the model is trained on the training data of these 24 classes. The 21 action classes from JHMDB-21 are used as novel classes. We evaluate the model on both the base classes and the novel classes. 

\noindent \bf OV-AVA. \rm OV-AVA is based on the AVA~\cite{ava} dataset.  AVA contains 211k video clips for training and 57k for validation. The video clips are segmented from 430 15-minute movie videos. There are 80 action classes in AVA. Following the standard protocol, the 20 classes with less than 25 validation examples are not used. The remaining 60 categories include 13 human poses, 32 human-object interactions, and 15 human-human interactions. Each human bounding box is annotated with one or more action classes, thus action detection on AVA is a multi-label detection task. As shown in Figure~\ref{ava}, we divide the 60 action classes into 45 base and 15 novel classes. The division is based on three principles: First, the ratio of the number of base classes and novel classes is 3:1, and the number of base classes and novel classes in each action type should also be close to 3:1. Second, the distribution of the number of instances of the base classes and the novel classes should be similar, and both follow the long-tail distribution. Third, the methods under the fully supervised setting should achieve similar performance on the base classes and novel classes.

\noindent \bf Evaluation Metric. \rm We take the frame-level mean average precision at 0.5 IoU threshold (Frame mAP@0.5) as the evaluation metric. A predicted action instance is recognized as a true positive if and only if its bounding box has an IoU overlap higher than a preset threshold with the ground-truth bounding box and the predicted action class matches the ground-truth action class. Following common practice, we set the IoU threshold to 0.5.
\section{Experiments}
\subsection{Implementation Details}

\noindent \bf Human Detector. \rm For UCF-JHMDB, we train a DINO~\cite{dino} detector on the video frames with action annotations of the UCF101-24 training set using the MMDetection~\cite{mmdetection} framework. The DINO detector is pretrained on the COCO~\cite{coco} object detection dataset. As the UCF101-24 dataset has a relatively small scale, we supplement the predicted bounding boxes into the training set to improve the robustness of the model. For validation, we use all human boxes with confidence scores higher than 0.3 as human proposals. For OV-AVA, following common practice, the human bounding boxes provided by LFB~\cite{lfb} are adopted. The human boxes are detected by a Faster R-CNN~\cite{fasterrcnn} detector which is pretrained on ImageNet and the COCO human keypoint detection dataset and fine-tuned on the keyframes of the AVA training set. We use ground-truth human boxes for training and all human boxes with confidence scores higher than 0.8 as human proposals for validation. 

\noindent \bf Video and Text Encoders. \rm We use the ViCLIP~\cite{wang2023internvid} models pretrained on the InternVid-10M-FLT dataset for video and text feature extraction. Both video encoder and text encoder use ViT structure. Unless otherwise stated, we use ViT-B/16 models. The expected resolution of the input videos is $224 \times 224$ and when the resolution of the input video is not this size, we perform interpolation on the positional embedding. The input video clips contain 8 frames. The temporal sampling stride is 1 and 8 for UCF-JHMDB and OV-AVA respectively.   

\noindent \bf Training and Inference Recipes. \rm The model is trained and evaluated on 8 NVIDIA RTX A5000 GPUs. For all training processes, we use an SGD optimizer with momentum 0.9 and weight decay $1 \times 10^{-7}$. For video region-text alignment pretraining, the batch size is set to 16. The model is trained for 60000 iterations with a base learning rate of $1 \times 10^{-5}$. For finetuning on base classes of UCF-JHMDB, the batch size is set to 32. The model is trained for 70000 iterations with a base learning rate of $5 \times 10^{-5}$. For finetuning on base classes of OV-AVA, the batch size is set to 8. The model is trained for 140000 iterations with a base learning rate of $1 \times 10^{-4}$. Random scaling, random horizontal flipping, and color jittering are adopted for training data augmentation. For inference, we scale the short size of input frames to 256 following common practice.

\begin{table}[t]
\centering
\scriptsize
\setlength{\tabcolsep}{9pt}
\caption{\textbf{Ablations on the global and local video features fusion ratio $\beta$.} Setting $\beta$ to 1 means only using local region features. We set $\beta$ to 0.3 as the default choice for UCF-JHMDB.}
\label{tab:abbeta}
\begin{tabular}{cl|ccccccc}
\toprule
\multicolumn{1}{l}{} & Fusion Ratio $\beta$ & 0 & 0.1 & 0.3 & 0.5 & 0.7 & 0.9 & 1 \\ \midrule
\multirow{2}{*}{mAP} & Base & 69.31 & 71.48 & 71.93 & 70.91 & 70.41 & 69.64 & 70.03 \\
 & Novel & 56.99 & 57.01 & 57.19 & 56.23 & 53.74 & 48.79 & 45.67  \\ \bottomrule
\end{tabular}
\vspace{-16pt}
\end{table}

\subsection{Ablation Study}

We conduct ablation experiments on the UCF-JHMDB dataset to investigate the influence of each design in our OV-STAD framework. We report Frame mAP on both base classes and novel classes for performance comparisons. 

\noindent \bf Ablations on Global and Local Video Feature Fusion. \rm We first investigate the value of $\beta$ which controls the proportion of local and global information in the base class finetuning process in Table~\ref{tab:abbeta}. Video region-text alignment is not adopted for these ablations to reduce the computational burden. On UCF-JHMDB, setting $\beta$ to 0.3 achieves the best results. This is because the UCF101-24 dataset has a relatively small scale and tends to overfit. Introducing global video features trained by alignment with texts at scale can alleviate this overfitting and improve the performance on novel classes. At the same time, the recognition of some action classes in UCF101-24 and JHMDB-21 relies on the perception of the global context. In Table~\ref{tab:ab1}, we further investigate the effect of global and local video feature fusion in both video region-text alignment and base class finetuning. Global and local feature fusion constantly brings performance improvement on novel classes.

\noindent \bf Ablations on Video Region-Text Alignment. \rm In Table~\ref{tab:ab1}, we investigate the influence of video region-text alignment pretraining. The first line is the result of ViCLIP directly performing zero-shot inference on both base classes and novel classes. We use the global video feature for each box to calculate cosine similarity with the embeddings of action category prompts. This is unreasonable because different human boxes in the video clip are classified using exactly the same features. However, when we extract RoI features for each human box to align with the action prompt embeddings, the detection mAP on the base classes and novel classes dropped to 5.08 and 12.52 respectively. This shows that the video-text pretrained model is indeed deficient in region-text alignment. From rows 4 and 5 in Table~\ref{tab:ab1}, it can be inferred that video region-text alignment pretraining effectively enhances video region features to make them better aligned with action prompt embedding. At the same time, when training on the video region-text pairs, the model can be exposed to dense descriptions of human actions and activities, thereby improving the ability to understand and distinguish various action concepts. After video region-text alignment pretraining, the detection performance of the model on base classes and novel classes is significantly improved, despite no spatio-temporal detection data being used for training. By comparing row 3 with row 8, we can find that video region-text alignment pretraining can still significantly improve the performance on the novel classes when base class fine-tuning is performed, which demonstrates that the model has learned richer action semantics. According to rows 3-4 and 8-9 in Table~\ref{tab:ab1}, local region features after video region-text alignment play a greater role in action recognition, bringing more performance gains. This further demonstrates the effectiveness of video region-text alignment training in enhancing local video region features.

\noindent \bf Ablations on Video Region-Text Alignment Training Data. \rm We investigate the influence of video region-text alignment training data in Table~\ref{tab:abdata}. Using only Video region-text pairs from HC-STVG2.0~\cite{hcvg} or only from VidSTG does not work as well as using Video region-text pairs from both datasets. However, using data only from HC-STVG2.0 works better than just using VidSTG~\cite{vidstg}, even though the amount of data from HC-STVG2.0 is smaller. This may be because the text description on HC-STVG2.0 involves more diverse action types, while the text description in VidSTG is annotated based on the VidOR~\cite{vidor} visual relation detection dataset and covers relatively limited actions. We further test using different proportions of video region-text pairs from both datasets. The experimental results show that the increase in video region-text alignment training data can improve performance.

\begin{table}[t]
\centering
\scriptsize
\vspace{-6pt}
\caption{\textbf{Ablation study on the training process.} ``VR-T Alignment'' denotes video region-text alignment. ``GL Fusion'' denotes using region features after fusion with global video features to align with text embeddings. The results show the effectiveness of video region-text alignment and base class finetuning. Global-local fusion in both processes can bring performance improvements. Default choice for our model is colored in \colorbox[HTML]{EFEFEF}{gray}.  }
\label{tab:ab1}
\setlength{\tabcolsep}{15pt}
\begin{tabular}{cc|cc|cc}
\toprule
\multicolumn{2}{c|}{ VR-T Alignment} & \multicolumn{2}{c|}{Base Class Finetuning} & \multicolumn{2}{c}{mAP} \\
\cline{1-6}
Local Only & GL Fusion & Local Only & GL Fusion & Base & Novel \\ \midrule
- & - & - & - & 13.55 & 26.02 \\
- & - & \checkmark & - & 70.03 & 45.67 \\
- & - & - & \checkmark & 71.93 & 57.19 \\
\checkmark & - & - & - & 32.47 & 50.46 \\
- & \checkmark & - & - & 45.06 & 60.67 \\
\checkmark & - & \checkmark & - & 71.62 & 54.88 \\
\checkmark & - & - & \checkmark & 71.86 & 60.81 \\
\rowcolor[HTML]{EFEFEF} 
- & \checkmark & - & \checkmark & 71.63 & 61.41 \\ \bottomrule
\end{tabular}
\vspace{-12pt}
\end{table}

\begin{table}[t]
\centering
\scriptsize
\setlength{\tabcolsep}{8pt}
\caption{\textbf{Ablations on the video region-text alignment pretraining data.} We test using video region-text pairs from HC-STVG2.0 or VidSTG only, as well as partial data from both datasets.}
\label{tab:abdata}
\begin{tabular}{cl|ccccccc}
\toprule
\multicolumn{1}{l}{} & VR-T Data & None & HC-STVG2.0 & VidSTG & 25\% & 50\% & 75\% & ALL \\ \midrule
\multirow{2}{*}{mAP} & Base & 71.93 & 71.81 & 71.48 & 71.04 & 71.43 & 71.65 & 71.63 \\
 & Novel & 57.19 & 60.08 & 59.30 & 58.43 & 60.12 & 60.98 & 61.41 \\ \bottomrule
\end{tabular}
\vspace{-16pt}
\end{table}

\noindent \bf Ablations on Base Class Finetuning. \rm As shown in rows 5 and 8 in Table~\ref{tab:ab1}, finetuning the model on base classes for the target action detection task can significantly improve the performance on base classes. The performance on novel classes is also improved as the model can be exposed to real action prompt texts and further learn to distinguish between actions.

\begin{table}[t]
\centering
\setlength{\tabcolsep}{14pt}
\scriptsize
\caption{\textbf{Results On UCF-JHMDB.} \checkmark of column ``Flow" denotes optical flow inputs are used. Zero-shot results of VLMs are tested by us.}
\label{tab:ucf}
\begin{tabular}{lccccc}
\toprule
 &  &  &  & \multicolumn{2}{c}{Frame mAP@0.5} \\ \cmidrule(l){5-6} 
\multirow{-2}{*}{Method} & \multirow{-2}{*}{Backbone} & \multirow{-2}{*}{Input} & \multirow{-2}{*}{Flow} & Base (UCF101-24) & Novel (JHMDB-21) \\ \midrule
\multicolumn{6}{l}{\cellcolor[HTML]{FFFFFF}Fully-supervised methods} \\ \midrule
Peng \etal~\cite{peng2016multi} & VGG & 1$\times$1 & \checkmark & 39.9 & 58.5 \\
ACT~\cite{ACT} & VGG & 6$\times$1 & \checkmark & 69.5 & 65.7 \\
TACNet~\cite{tacnet} & VGG & 16$\times$1 & \checkmark & 72.1 & 65.5 \\
ACRN~\cite{acrn} & S3D-G & 20$\times$1 & \checkmark & - & 77.9 \\
AVA~\cite{ava} & I3D & 20$\times$1 & \checkmark & 76.3 & 73.3 \\
MOC~\cite{li2020actions} & DLA34 & 7$\times$1 & \checkmark & 78.0 & 70.8 \\
MOC~\cite{li2020actions} & DLA-34 & 7$\times$1 & \checkmark & 73.1 & - \\
T-CNN~\cite{t-cnn} & C3D & 7$\times$1 & \checkmark & 41.4 & 61.3 \\ \midrule
\multicolumn{6}{l}{\cellcolor[HTML]{FFFFFF}Zero-shot results of VLMs} \\ \midrule
Open-VCLIP~\cite{weng2023open} & ViT-B/16 & 8$\times$8 & \xmark & 5.2 & 7.0 \\
VideoCLIP~\cite{xu2021videoclip} & S3D & 8$\times$8 & \xmark & 6.9 & 5.3   \\
ViCLIP~\cite{wang2023internvid} & ViT-B/16 & 8$\times$8 & \xmark & 13.6 & 26.0 \\ \midrule
Ours & ViT-B/16 & 8$\times$1 & \xmark & 71.6 & 61.4 \\ \bottomrule
\end{tabular}
\vspace{-14pt}
\end{table}

\subsection{Quantitative Results}
We provide the OV-STAD results of the proposed method on the UCF-JHMDB dataset in Table~\ref{tab:ucf}. We first compare with some fully supervised methods. Our method uses label supervision on the base classes, achieving comparable results with the previous method. Without using any label supervision of the novel classes, our method achieves a Frame AP@0.5 of 61.4 on the novel classes, which is close to the result of T-CNN~\cite{t-cnn} under full supervision (61.4 vs 61.3). The performance gaps with the strong ACT~\cite{ACT} and MOC~\cite{li2020actions} detectors are only 4.3 and 9.4 respectively, noting that they use extra optical flow input. We also compare with zero-shot results of some VLMs. These methods perform well on zero-shot video classification tasks but have low accuracy on fine-grained spatio-temporal action detection tasks. This is because there is a gap between the upstream pretraining and the target action detection task. We choose ViCLIP~\cite{wang2023internvid}, which performs relatively well, as the baseline, and uses video region-text alignment pretraining and global and local visual feature fusion to reduce the gap between pretraining and the downstream task. 

We also provide the results on OV-AVA in Table~\ref{tab:ava}. Fully supervised methods have made great progress on AVA, benefiting from large-scale action classification dataset pretraining and advances in spatio-temporal context modeling technology such as the introduction of long-term memory~\cite{aia,acarn,TubeR,wu2023stmixer}. However, the zero-shot results of VLMs on AVA are unsatisfactory. This is because the action classes of AVA are all atomic actions, without specific related objects and scenes objects and scenes involved. In the video-text pre-training process, the video-text alignment learned by the model relies heavily on object appearance and scene cues. There are a large number of multi-person scenes in AVA. Different people in the same scene are often performing different actions, which puts higher requirements on region-text alignment. Our method achieves a Frame mAP@0.5 of 16.6 on the novel classes of OV-AVA, comparable to some early fully supervised results.

\begin{table}[t]
\centering
\scriptsize
\setlength{\tabcolsep}{13pt}
\caption{\textbf{Results on OV-AVA.} The methods marked with * use AVA v2.1 annotations, which affects the detection mAP slightly.}
\label{tab:ava}
\begin{tabular}{lcccccc}
\toprule
 &  &  &  & \multicolumn{3}{c}{Frame mAP@0.5} \\ \cmidrule(l){5-7} 
\multirow{-2}{*}{Method} & \multirow{-2}{*}{Pretrain} & \multirow{-2}{*}{Backbone} & \multirow{-2}{*}{Input} & Base & Novel & All \\ \midrule
\cellcolor[HTML]{FFFFFF}Fully-supervised methods &  &  &  &  &  &  \\ \midrule
SlowFast~\cite{feichtenhofer2019slowfast} & K400 & Slowonly-R50 & 4$\times$16 & - &-& 19.2 \\
SlowFast~\cite{feichtenhofer2019slowfast} & K600 & SF-R101-NL & 32$\times$2 & - & - & 29.0 \\
VTr*~\cite{vat} & K400 & I3D & 64$\times$1 & - & - & 24.9 \\
AVA*~\cite{ava} & K400 & I3D & 20$\times$1 & - & - & 14.6 \\
ACRN*~\cite{acrn} & K400 & S3D-G & 20$\times$1 & - & - & 17.4 \\
STEP*~\cite{STEP} & K400 & I3D & 12$\times$1 & - & - & 18.6 \\
ACARN~\cite{acarn} & K600 & SF-R101-NL & 32$\times$2 & - & - & 31.4 \\
TubeR~\cite{zhao2022tuber} & IG-65M & CSN-152 & 32$\times$2 & - & - & 33.4 \\
WOO~\cite{chen2021watch} &  K600 & SF-R101-NL & 32$\times$2 & - & - & 28.3 \\
AIA~\cite{aia} & K700 & SF-R101 & 32$\times$2 & - & - & 32.3 \\
STMixer~\cite{wu2023stmixer} & K700 & SF-R101 & 32$\times$2 & - & - & 32.9 \\
STMixer~\cite{wu2023stmixer} & IG-65M & CSN-152 & 32$\times$2 & - & - & 34.8 \\ \midrule
\cellcolor[HTML]{FFFFFF}Zero-shot results of VLMs &  &  &  &  &  &  \\ \midrule
VideoCLIP~\cite{xu2021videoclip} & HowTo100M & S3D & 8$\times$8 & 3.9 & 5.2 & 4.2 \\
Open-VCLIP~\cite{weng2023open} & K400 & ViT-B/16 & 8$\times$8 & 3.7 & 5.8 & 4.2 \\
ViCLIP~\cite{wang2023internvid} & InternVid & ViT-B/16 & 8$\times$8 & 4.3 & 5.3 & 4.6 \\ \midrule
Ours & InternVid & ViT-B/16 & 8$\times$8 & 27.8 & 16.6 & 25.0 \\ \bottomrule
\end{tabular}
\vspace{-15pt}
\end{table}

\section{Conclusion}
In this paper, we have presented an OV-STAD framework. We propose two benchmarks for the OV-STAD task based on the existing STAD datasets. A simple but effective OV-STAD method is proposed based on VLMs. The core design is to further pretrain the model on video region-text pairs with high-density human action descriptions to enhance the alignment between the video regions and texts as well as improve the understanding of diverse action concepts. Global video feature and local region feature fusion is applied before aligning with text embeddings to retain and leverage existing knowledge from video-text pretraining and provide a global context. Our method achieves promising detection results on two benchmarks. We hope this paper can provide a solid baseline for future OV-STAD research.

{
\small
\bibliographystyle{ieee_fullname}
\bibliography{main}
}

\end{document}